\documentclass{article}

\usepackage{booktabs} 
\usepackage{graphicx} 
\usepackage{xcolor}
\PassOptionsToPackage{numbers, compress}{natbib}
\usepackage{natbib}
\usepackage[utf8]{inputenc}
\usepackage[a4paper, total={6in, 10in}]{geometry}
\usepackage{graphicx}
\usepackage{indentfirst}
\usepackage{authblk}
\usepackage{lineno}
\usepackage{comment}
\usepackage{graphicx}
\usepackage{subcaption}
\usepackage{comment}
\usepackage{url}

\title{Demographic Predictability in 3D CT Foundation Embeddings}
\author[1]{Guangyao Zheng}
\author[1,2,3]{Michael A. Jacobs}
\author[4]{Vishwa S. Parekh}
\affil[1]{Department of Computer Science, Rice University, Houston, TX, USA}
\affil[2]{Department Of Diagnostic And Interventional Imaging, McGovern Medical School, UTHealth Houston, Houston, TX, USA}
\affil[3]{Department of Radiology and Radiological Science and Sidney Kimmel Cancer Center, Johns Hopkins University School of Medicine, Baltimore, MD 21205, USA}
\affil[4]{University of Maryland Medical Intelligent Imaging (UM2ii) and Department of Diagnostic Radiology and Nuclear Medicine, University of Maryland School of Medicine, Baltimore, MD 21201, USA}

\begin{document}

\maketitle

\section{Introduction}
Self-supervised learning has made substantial progress in medical imaging by enabling efficient and generalizable feature extraction from large-scale unlabeled datasets. Self-supervised foundation models have recently been successfully extended to encode three-dimensional (3D) computed tomography (CT) data into computation-efficient, information-rich embedding with 
1408 features, with excellent performance across several downstream tasks such as intracranial hemorrhage detection and lung cancer risk forecasting \cite{yang2024advancing,Kiraly_Traverse_2024,CT_Foundation_Demo}. However, as self-supervised models learn from complex data distributions, questions arise concerning whether these embeddings capture demographic information, such as age, sex, or race, as they could have significant advantages (demographically-aware personalized clinical decision support systems) or disadvantages (exposure to potential compromise of the fairness of clinical applications \cite{gichoya2022ai, kulkarni2024hidden}). This letter addresses a preliminary investigation into whether self-supervised 3D CT embeddings encode demographic information.

\section{Materials and Methods}
In this retrospective study, we used The National Lung Screening Trial (NLST) public dataset, with 3D CT images of the lung from patients aged 55-74 with demographic information (age, sex, and race) \cite{national2011national} (Table 1(a)). The CT Foundation tool \cite{yang2024advancing,Kiraly_Traverse_2024,CT_Foundation_Demo} provides embeddings for the NLST dataset and patient-wise training (N=10299 patients, 52696 images) and test (N=2199 patients, 11421 images) data splits. We trained different models (softmax and linear regression, linear support vector machine, random forest, and decision tree) for predicting sex (N=2), race (N=3, following previous studies \cite{gichoya2022ai}), and age. The performance metrics for regression were root mean square error (RMSE) and mean absolute error (MAE). The performance metrics for classification were group-wise accuracy and AUC (Area under the curve) of the ROC (Receiver operating characteristic). Statistical significance was set at $p < 0.05$. Our code is available at \url{https://github.com/BioIntelligence-Lab/CT-Embedding-Bias}.

\begin{table}[!htb]
\center
\includegraphics[width=1\textwidth]{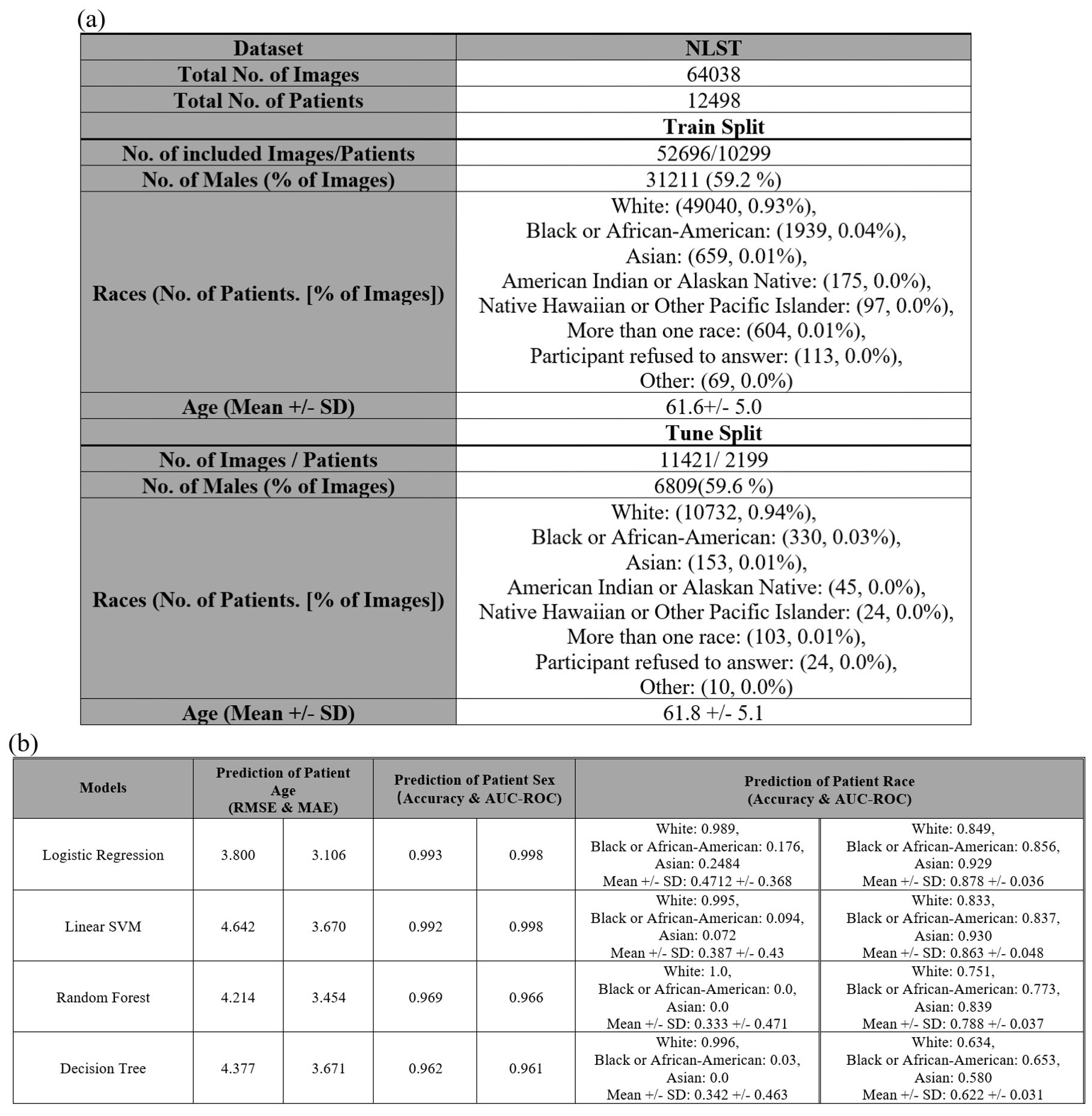}
\caption{(a) Detailed statistics of the NLST dataset and (b) the model performances on patient demographics} \label{fig1}
\end{table}

\section{Results}
The models trained using CT Foundation embeddings accurately predicted age and sex information but not race information. The linear regression model had the best performance and predicted age with an RMSE of 3.8 years, while the softmax regression model had the best classification performance, predicting sex and race with an AUC of 0.998 and 0.878, respectively. The accuracy scores for sex and race were 0.993 and 0.471, respectively. The detailed performance report is shown in Table 1(b). The scatterplot illustrating the predicted vs. actual age, the t-SNE and Isomap plots are the 2D representation of the embeddings overlaid with sex classes, and the ROC curves for sex and race classification for the softmax regression model are illustrated in Figure 1. 

\begin{figure*}[!htb]
\center
\includegraphics[width=1\textwidth]{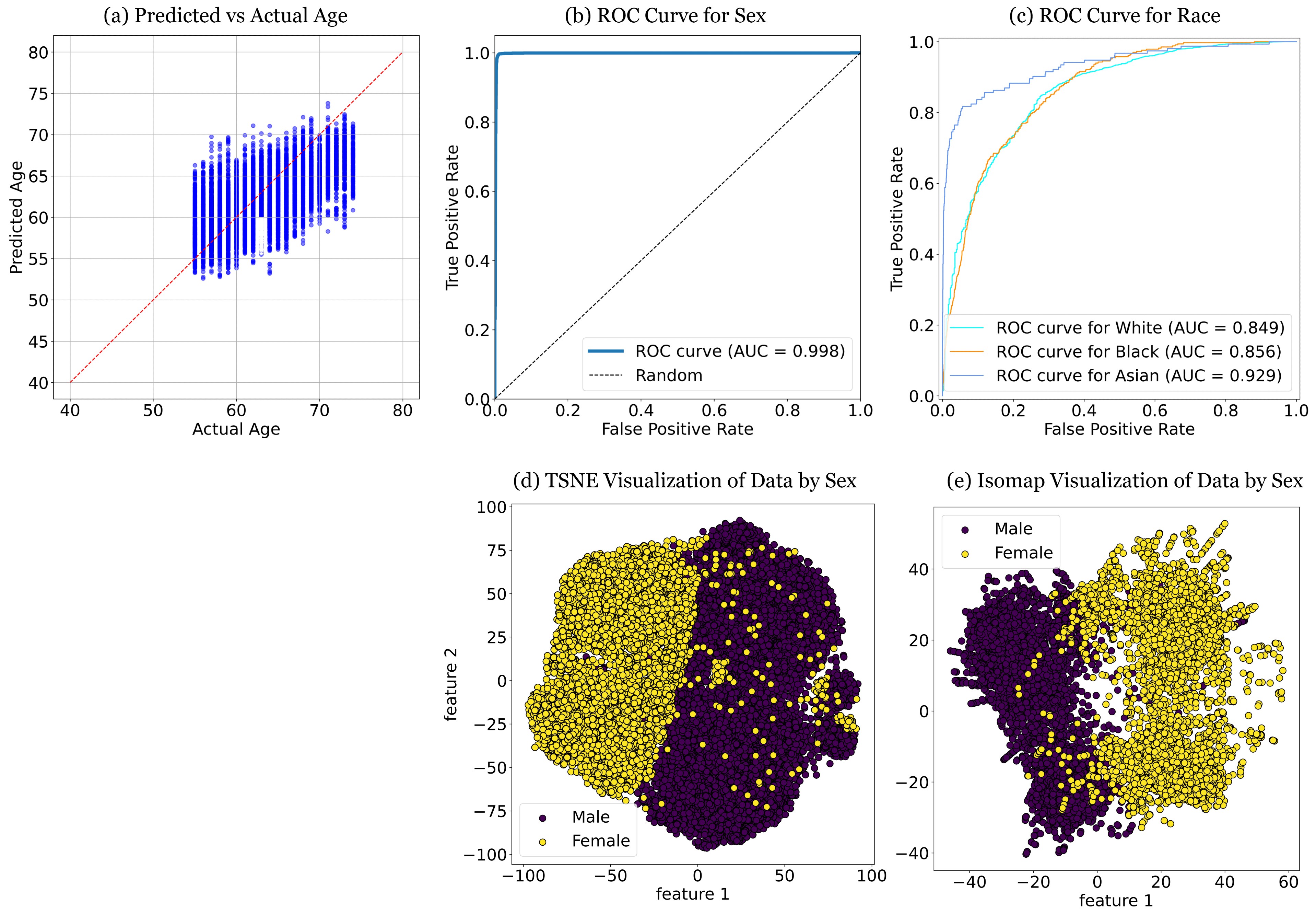}
\caption{(a) Logistic regression models predicted age vs. actual age compared. (b) ROC curve of the logistic regression model classification for sex. (c) ROC curve of the logistic regression model classification for race. (d) Visualization of the data colored by sex after T-SNE non-linear dimensionality reduction to two features.  (e) Visualization of the data colored by sex after Isomap non-linear dimensionality reduction to two features.} \label{fig2}
\end{figure*}

\section{Discussion}
Our results demonstrated that self-supervised CT embeddings can predict certain demographic features (sex and age) with excellent accuracy, indicating that the self-supervised CT embeddings indeed encode demographic characteristics. This information can either be useful for personalizing clinical decision support tailored to one’s sex and age or could potentially expose the model to demographic bias propagation and security vulnerability risks. In both cases, our work highlights the importance of understanding what information is being encoded within foundation model encodings to ensure downstream clinical applications’ safe and optimal development.

Our results indicate that the CT foundation model embeddings encoded sex and age more effectively than race. This may be because age and sex are biological characteristics with observable anatomical characteristics on a chest CT scan, while race lacks such direct biological representation in medical imaging. However, the race data in our study was predominantly categorized as “White” with $94\%$ of the patients, which reduced the efficacy of our models in effectively evaluating the presence of race information in the embeddings. Nevertheless, our models show that sex and age are encoded in the embeddings.

In conclusion, as self-supervised approaches gain traction in radiology, it is essential to balance their advantages with strategies to mitigate any safety concerns. This was a preliminary study evaluating the ability of a single CT model to encode demographic information using embeddings from a single dataset. Continued exploration into understanding the information in foundation model embeddings will help ensure that AI in medical imaging progresses responsibly, protecting patient privacy and enhancing fairness.

\bibliographystyle{unsrtnat}
\bibliography{reference}
\end{document}